# Effective Enabling of Sharing and Reuse of Knowledge On Semantic Web by Ontology in Date Fruit Model.

Sherimon. P.C [1]    Reshmy Krishnan[2]    Vinu.P.V[3]

[1]Dept.Of IT, *Arab Open University, Sultanate of Oman*

[2]Dept.Of Computing, Muscat College

[3]Dept of IT, Higher College Of Technology

**Abstract**

Since Organizations have recognized that knowledge constitutes a valuable intangible asset for creating and sustaining competitive advantages, knowledge sharing has a vital role in present society. It is an activity through which information is exchanged among people through different media. Many problems face the area of knowledge sharing and knowledge reuse. Currently, knowledge sharing between entities is achieved in a very ad-hoc fashion, lacking proper understanding of the meaning of the data. Ontologies can potentially solve these problems by facilitating knowledge sharing and reuse through formal and real-world semantics. Ontologies, through formal semantics, are machine-understandable. A computer can process data, annotated with references to ontologies, and through the knowledge encapsulated in the ontology, deduce facts from the original data. The date fruit is the most enduring symbol of the Sultanate's rich heritage. Creating ontology for dates will enrich the farming group and research scholars in the agro farm area.

## 1. Introduction

Many problems face the area of knowledge sharing and knowledge reuse. Currently, knowledge sharing between entities is achieved in a very ad-hoc fashion, lacking a proper understanding of the meaning of the data. Semantic search seeks to improve search accuracy by understanding searcher intent and the contextual meaning of terms as they appear in the searchable data space, whether on the Web or within a closed system, to generate more relevant results. Semantic Search uses semantics or the science of meaning in language, to produce highly relevant search results [1]. In most cases, the goal is to deliver the information queried by a user rather than have a user sort through a list of loosely related keyword results. Structured data sources like ontology act as the backbone to Semantic web for achieving such highly related results. Such technologies enable the formal articulation of domain knowledge at a high level of expressiveness and could enable the user to specify his intent in more detail at query time. Ontology is the backbone of semantic web and it means a specification of a conceptualization. That is, ontology is a description (like a formal specification of a program) of the concepts and relationships that can exist for an agent or a community of agents. This definition is consistent with the usage of ontology as set-of-concept-definitions.

## 2. Why to develop Ontology?

Ontology defines a common vocabulary for researchers who need to share information in a domain. It includes machine-interpretable definitions of basic concepts in the domain and relations among them.

They are used as a unified framework or shared understanding of communication between people and systems [3].Sharing of the Information, reusing and updating of the existing ontology is the major merits of the Ontology development [2].By developing Ontology we can share the common understanding of the structure of information among people or software agents. Different websites may contain variety of details about the date fruit. Through Ontology all these information can be aggregated and can be published .The users and agents can use this aggregated information to answer user queries or as input data to other applications. The Developed ontology can be reused in the future for other users. If they need to build a large ontology, the existing ontologies describing portions of the large domain can be integrated .The existing ontology can be updated also.

## 3. Ontology in knowledge sharing

Ontologies have the potential of enabling true knowledge sharing and reuse among heterogeneous agents, both human and computer. Because there are so many different types of ontologies, ranging from simple word lists to comprehensive ontologies with the expressive power of full first-order logic, we can depend on ontology for any type of knowledge representation. The expressiveness of ontology relates to the degree of explication of the (meta-) knowledge, which is captured in the ontology. When more relations and more constraints are captured in the ontology, the ontology becomes more expressive, since it captures the knowledge of the domain on a more detailed level.

**4. Date Fruit Taxonomy and Property Definition.**

The date fruit is the most enduring symbol of the Sultanate's rich heritage, along with other cherished aspects of traditional Omani life. It has been the main wealth of people in past generations, the fruit serving as a source of daily nourishment, Phoenix dactylifera is the true Date fruit from which the tasty fruit is obtained. With a history stretching back over 5,000 years, this venerable fruit grows in thick clusters on the giant Date fruit, native to the Middle East. Dates require a hot, dry climate and, besides Africa and the Middle East, flourish in California and Arizona. 60% of worlds date production is from Arab countries.

Date fruit has only one seed, and can vary in size, shape, color and quality of flesh. Unripe Dates are green in color, maturing to yellow, then reddish-brown when fully ripe. Wild Dates are morphologically and ecologically similar to domesticated Dates but have smaller, inedible fruits. The maximum length of the fruit is about 6cm. All Dates have a single, long, narrow seed. The skin is thin and papery, the flesh cloyingly sweet.

Creating taxonomy for dates will enrich the farming group as well as a model for the research scholars in the agro farm area. It is not always possible to build an all comprehensible ontology in a single stretch. It requires several iteration through several well equipped persons. We tried to create ontology of dates in minimum requirement basis. Different aspects of date fruit such as attributes, benefits, chain of operation from harvest delivery, developing stages, quality profile, composition and products are being described in the taxonomy.

4.1. Attributes

Attributes class covers the color, shape, size, taste and texture of the date fruits.

4.2 Benefits

This section describes the benefits of date fruit in Food and health.
1) Food: Dates has a leading role on the dining table of Middle East. A good quality date drupe is a delicious fruit with a sweet taste and a fleshy mouth feel. This is a high-energy food containing sugars and fiber thus being suitable for both people and livestock. The applications of dates in different food types are listed out such as condiments, deserts, etc. Date paste can be used in mixtures and bakery products. Dates can be used as a good preservative also. Whole pitted dates are also a delicious type of food.

2) Health: Health benefits of dates are uncountable, as this fruit is affluent in natural fibers. Dates are even rich in several vitamins and minerals. These natural products contain oil, calcium, sulphur, iron, potassium, phosphorous, manganese, copper and magnesium which are advantageous for health. It is said that consumption of one date daily is necessary for a balanced and healthy diet. Dates help in fighting constipation, intestinal disorders, weight gain, heart problems, diarrhea and abdominal cancer.

4.3 Chain of operation from harvest & delivery

This class describes various operations done for preparing dates from harvest to delivery. It includes transport of dates from the harvesting fields, additional treatments required, packing, sorting and cleaning and storage methods. Additional treatments consist of coating, dehydration, glazing, hydration, maturation and pitting. Different techniques to prevent infect infestation during storage are desribed.They are fumigation, heat treatment, irradiation and refrigeration.

4.4. Developing stages

The edible stages of ripening of date fruit could be divided into five phases .They are hababauk,khalaal,Kimri,Rotab and tamr. One week fruit is called Hababuk.Fruit color is yellow in khalaal stage, Bright brown in rotab and then becomes dark brown in tamr stage. Even though date is edible in all three phases, rotab is the most delicious. At khalaal stage weight gain is slow but sucrose content increases, moisture content goes down, and tannins will start to participate and lose their astringency. In rotab stage the tips of the fruit starting to turn brown and weight of the fruit will be reduced due to moisture loss. Softening of tissues, browning of the skin is also happening during this stage. When the they enter to tamr, the dates are left to ripen further on the palm [15].

4.5. Quality Profile

The quality of the date fruits is determined based on the factors such as attributes, composition, defects and presence of other particles. Different defects which are frequently counted are blemishes, broken skin, deformity, discoloration, shrivel and sunburn. In the quality checking presence of other particles will be considered. Mainly checking will be for foreign matters, insect infestation and pesticide residue.

4.6 Composition

The composition of date fruit are enzymes, vitamins, minerals, crude fibers, moisture, proteins, fats, sugars and chemical substances. Main chemical substances are organic acids, and polyphenols.

4.7. Products

Various products of dates like are listed out in this section. Some of the delicious products are date condiments, date deserts, date paste, date preserves, whole pitted dates. Date paste is used in bakery products, in mixture and in pure date paste.

**5. Date Fruit Ontology and Implementation**

With the enormous application of semantic web, ontologies are become more widely available. There is no single standard way to develop Ontology to define ontology. It is not necessary to start from scratch always. If there is an ontology existing in that domain by a third party, it will give a good starting for our ontology. Updating is also possible to the existing ontology. The date fruit ontology referred pepper ontology.

The development of Ontology depends on the needs and expectations of the users of different services. We have to develop ontology based on the applications that you have in mind and the extensions that you anticipate. Ontologies are able to specify different kinds of concepts such as classes, relationships between things and properties of things. According to Noy and McGuinness [], the following phases are to be followed to develop ontology.

5.1. Determine the domain and scope of the ontology.

Developing ontology without any purpose is not a goal itself. Ontology is a model of a particular domain built for a particular purpose. In the same domain itself ontology will vary according to the purpose and type of query aims. Ontology is by necessity an abstraction of the construction of ontology determined by the use to which the ontology will put and by future extensions that are already anticipated. Basic questions to be answered at this stage are what the domain is and the ontology will cover? For what we are going to use the ontology? For what type of queries should the ontology provide answers? Who will be users of this ontology? We have considered the following questions in the construction of ontology.

- What are the attributes of date fruit?
- What are the benefits of date fruit?
- What are the major operations to prepare dates?
- What are the developing stages of dates?
- How to check the quality of dates?
- What are the different products of dates?
- What are the compositions of a date fruit?

Since the date fruit is considered the most important crop in Sultanate of Oman and has a significant presence among the Oman society and it is valued for its social, religious and agricultural value, we have identified the domain as dates. Fruits. We hope this will enrich the farming group as well as the model for the research scholars in the agro-farm area.

5.2. Consider reuse.

Ontology has become widely available with the spreading of semantic web. It's always better to check for the particular ontology in the chosen domain. Rarely we have to start from the scratch. There is almost always an ontology available from a third party that provides a useful starting point of our ontology. There are so many libraries of reusable ontologies on the web such as Ontolingua Ontology library, DAML ontology library etc. Since the ontology for date fruit is not available in the libraries so far, we have developed from scratch.

5.3. List Key terms

All the terms that are likely to appear in the ontology can be listed out. The relations among the classes and the properties of the classes and instances in the ontology also to be listed out. As per our domain, important terms include date fruit, dates, quality profile, compositions developing stages, etc.

5.4 Define taxonomy

After the identification of key terms, these terms must be organized in a taxonomic hierarchy. There are different approaches to define classes such as top-down, bottom-up and combined one. Whatever be the method following, it is important to ensure that the hierarchy is a taxonomic hierarchy. That is If B is a subclass of A, then every instance of B must also be an instance of A. Only this will ensure that we follow the built in semantics of primitives such as owl: subclass Of and rdfs: subClassOf.In the Date fruit Ontology we follow the top-down approach. Date fruit is the main root class. Subclasses of Date fruit are Dates, Products and species. Under these subclasses, more subclasses are being defined.

5.5. Define the properties of the classes.

Properties define the relationships between two objects. There are two types of properties. Object properties and data properties. Object properties are used to link object to objects. Data Properties are used to link objects to xmlschema data type. OWL has another property – Annotation properties, to be used to add annotation information to classes, individuals, and properties.

Different object properties such as has_ deciding factor, has_composition, has_benefits etc. are used in the ontology. Properties can be allotted to a class. All subclasses of a class inherit the property of that class. If we apply the property, has_ features to the class dates, the same will be applicable for all subclasses of the class dates. Additional properties can be added to subclasses also.

Data properties such as has_ common name, has _country_of_origin, has_date_of_origin etc. is added in the ontology. They are used to link the instance and a class. For giving a value to an instance in a class we use data properties. For example the common name for 'Barhee' date is 'honey balls'. The data property is applicable to each instance of a class. It can't be apply to a general class.

5.6. Define the facets of a class

Facets of a property describe the value type, allowed values, the cardinality and other features of the values the property can take. For example consider the property has _common name. The value type of this is 'string'.

1) Value type: This describes the different types of values a property can take. The property has_date_of_origin has the value type Number. Other types include String, Boolean, Enumerated, Datetime, literal etc.

2) Allowed values: This represents values allowed for different properties. The property has_common_name has allowed values are "honey balls' and "visitors dates" etc.

3) Cardinality: A property can have single value or multiple values. Cardinality defines how many values a property can have. If the system allows 'at most a single value', it is single cardinality. Cardinality is multiple if it allows 'at least one value'.

5.7 Define instances

Individual instances of the classes are created in the ontology. We use ontology to organize sets of instances. Since the number of instances in ontology is quiet a large number when compared to the number of classes, creating instances is not done manually. Instead they are retrieved from legacy data sources such as databases. Automated extraction of instances from a text is also possible with the help of suitable plug-ins.

## 6. Implementation in Protégé 4.2

6.1 classes and subclasses.

Classes are the domain concepts and the building blocks of ontology. In Date fruit ontology, Date fruit is the subclass of OWL: Thing.

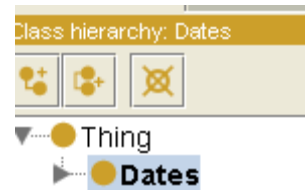

Fig.1.Top level Date fruit Taxonomy

A class can have subclasses which represents the middle level Taxonomy. The following taxonomy shows Dates has subclasses like Attributes, Benefits, Developing stages, Chain of operations, Quality profile composition and products.

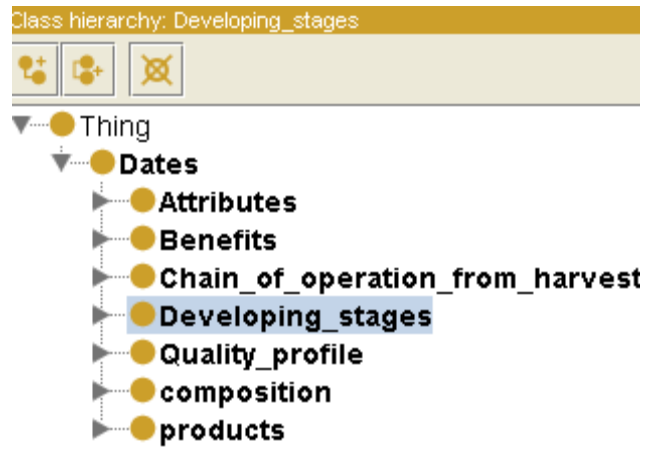

Fig.2.Middle level Date fruit Taxonomy

6.2. Properties, facets and instances.

In the Date fruit ontology, object property and data property has been defines. Individuals are also defined in the ontology. This will be created by choosing a class and then inserting the values of properties.

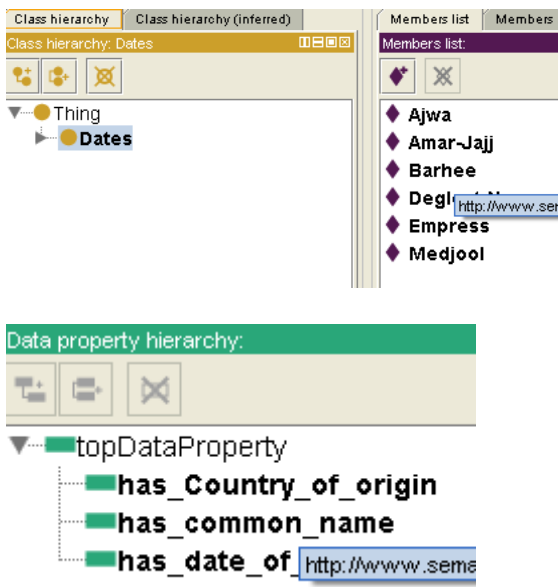

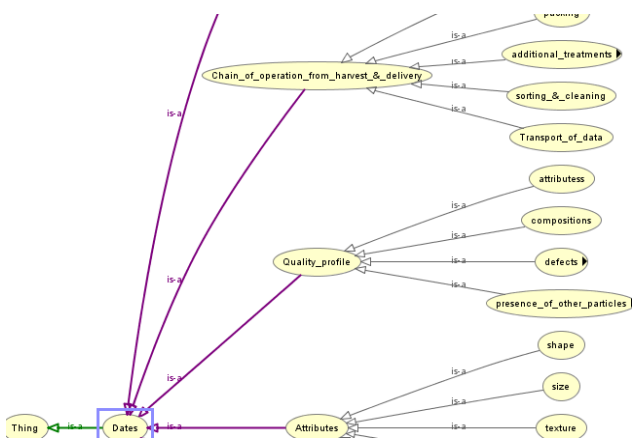

Fig.3.instances and data properties of date taxonomy

6.3. Class hierarchy in OWLVIz

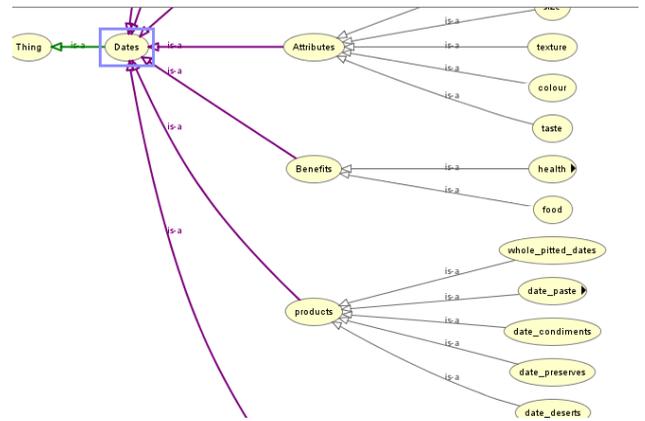

Fig.4.Example of class hierarchy

Using the plug in OWL VIz, the class hierarchy can be viewed .This allows comparison of asserted class hierarchy and inferred class hierarchy. The above figure shows the asserted model class hierarchy of Date fruit ontology.

6.4. Results

DL Query tab is used to create queries. The following fig shows the results of various queries given in the Query Tab.

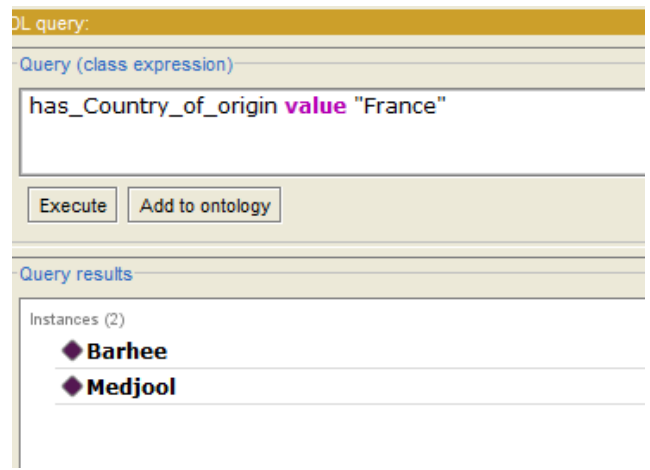

Fig.5.DL Query

## 7. Conclusions

One of the biggest bottlenecks for growth of information relates to the inability to share knowledge. The web has played an important part in the development of knowledge sharing. Currently, knowledge sharing between entities is achieved in a very ad-hoc fashion, lacking a proper understanding of the meaning of the data. Ontologies have the potential of enabling true knowledge sharing and reuse among heterogeneous agents, both human and computer .The expressiveness of ontology relates to the degree of explication of the (meta-) knowledge, which is captured in the ontology. When more relations and more constraints are captured in the ontology, the ontology becomes more expressive, since it captures the knowledge of the domain on a more detailed level. Different websites may contain variety of details about the date fruit. Through Ontology all these information can be aggregated and can be published .The users and agents can use this aggregated information to answer user queries or as input data to other applications. The Developed ontology can be reused in the future for other users. If they need to build a large ontology, the existing ontologies describing portions of the large domain can be integrated .The existing ontology can be updated also.

**Acknowledgments**

Steps for the construction of ontology has been taken ontology development 101: A guide to create your first ontology N.Noy, D.L.McGuinness

**Sherimon P.C.** holds a Master's Degree in Computer Applications from Madras University, India after finishing Bachelor in physics, pre degree and S.S.LC.He **is** a faculty of Computer Studies at Arab Open University (AOU), Muscat, Sultanate of Oman. His previous employment history lies in MG Unvty (India), Higher College of Technology (Muscat). His research interests lies in the area of semantic web and web programming. He has presented research papers in several National and International Conferences. He has 6 publications in International journals and book-"current computer world" in his account.

**Dr. Reshmy Krishnan** holds a PhD degree in computer science and engineering from Sathyabhama University, master's degree in computer science and engineering from Madras University India and bachelor degree from Madurai Kamaraj University. She is an Associate Professor at Muscat College, Muscat, and Sultanate of Oman. She is a reviewer in many journals and chaired conference sessions. Her research interests lies in the area of Ontology and Semantic web. She has presented several research papers in International Conferences. She has 13 publications in International journals. And 2 books.

**Vinu P.V.** is a lecturer in Information Technology at Higher College of Technology (HCT), Muscat, Sultanate of Oman. She holds a Master's Degree in Computer Science from Mahatma Gandhi University, India. Her research interests lies in the area of pervasive computing, web technologies and programming. She has presented research papers in various International Conferences. She has 6 publications in International journals.